# GENERATING SEGMENT DURATIONS IN A TEXT-TO-SPEECH SYSTEM: A HYBRID RULE-BASED/NEURAL NETWORK APPROACH

*G. Corrigan, N. Massey, and O. Karaali*
Speech Processing Laboratory
Motorola, Inc.
1301 E. Algonquin Rd., Schaumburg, IL   60196, U.S.A.

## ABSTRACT

A combination of a neural network with rule firing information from a rule-based system is used to generate segment durations for a text-to-speech system. The system shows a slight improvement in performance over a neural network system without the rule firing information. Synthesized speech using segment durations was accepted by listeners as having about the same quality as speech generated using segment durations extracted from natural speech.

## 1. INTRODUCTION

In any text-to-speech system, one of the problems is to establish the timing of the events in the speech signal. One common way to establish this timing is by setting segment durations, i.e., allotting time periods during which particular phones are nominally being uttered. One traditional method of determining these durations is to use a rule-based system [1], using rules based on the context in which the segment is set. These rules increase or decrease segment durations along a scale determined by the identity of the phone uttered during the segment. Each rule includes a "condition" element and an "action" element. The condition element determines whether the rule "fires", i.e., whether the action (some adjustment of a value used to compute the duration) is carried out. This method requires little computation and can incorporate the best available expert knowledge of the field into the model; however, the rules can be difficult to design, and represent a simplified model of segment duration, which may not be accurate.

An alternative method is to use a neural network to determine the segment durations [2]. The neural network is trained on a segmented database of speech, with the phonetic context of each segment provided as input, and its duration used as training data. The neural network then generates its own duration model, which can be more complex and extensive than that of the rule-based system. It should also be easier to train a network for a new language or speaking style than to develop a new rule set. One problem with this approach is that the size of the training set for the neural network is equal to the total number of phonetic segments labeled in the speech database. Especially with hand-labeled data, this limited data set may be too small to train a neural network. This can lead to the neural network overtraining on the available data, and not learning to generalize well to the overall speech domain. The neural network also requires more computation than the rule-based system.

This paper describes a neural network which incorporates input from a rule-based system to provide better performance on the problem of determining segment durations.

## 2. NEURAL NETWORK SYSTEM

Figure 1 illustrates the form of neural network system used to generate segment durations. For each phonetic segment, a vector is constructed describing the phone being uttered, the stress placed on the syllable containing the segment, the type of word containing the syllable, and whether the phone is the first or last phone in a syllable, word, phrase, clause, or sentence. The neural network is trained to map from a sequence of these vectors to the duration of the phonetic segment associated with the center vector. In all of the networks described here, the context window consists of a bit vector. The phones are identified in one portion of the vector using a one-of-n encoding to identify the phone, and in another portion as a vector of binary articulatory features. This redundant encoding does seem to enhance the performance of the network.

Increasing the width of the context window for the segment (i.e., the length of the sequence of vectors) increases the context information the network can use to determine the segment duration. On a small training database, however, a large context window allows the neural network to extract features that happen to correlate with the segment durations, but which wouldn't correlate across a larger database. This leads to a situation in which the errors found in training decrease while the errors found in testing the network on other data increase.

Reducing the size of the explicitly included phonetic context and replacing the lost data with a description of the important characteristics of that phonetic context can reduce this overtraining problem. The problem then becomes an issue of determining what characteristics of the removed phonetic context to use as input.

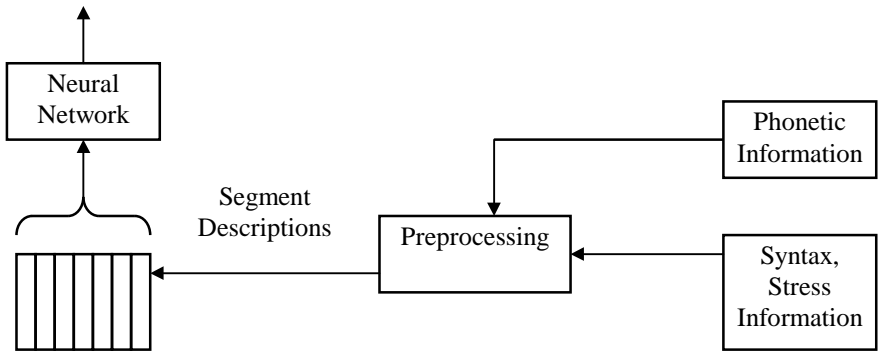

**Figure 1:** Neural Network for Segment Duration Computation

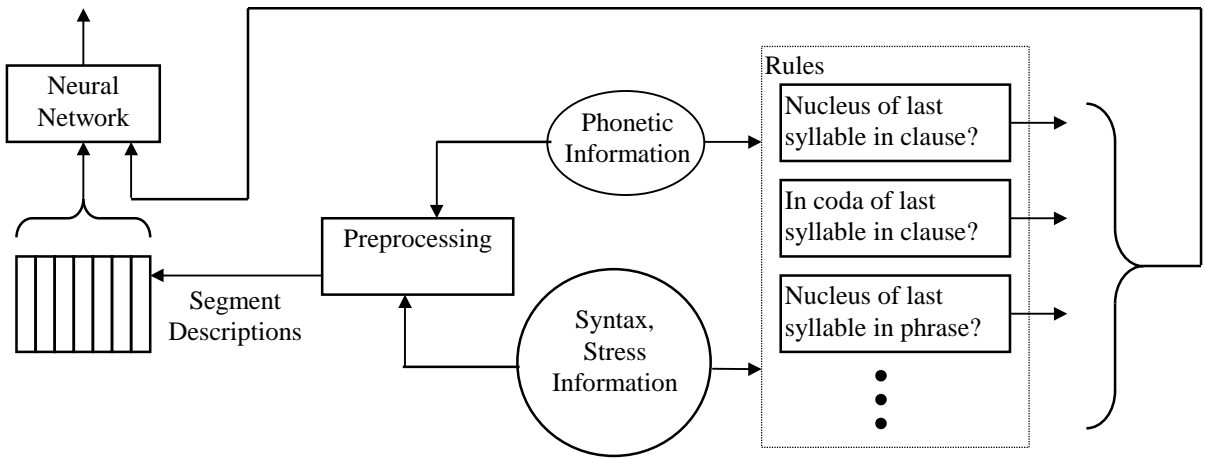

**Figure 2:** Hybrid System for Segment Duration Computation

Figure 2 illustrates a hybrid approach which combines a neural network with the condition elements of a rule-based system, i.e., the neural network input includes information about which rules would fire in the rule-based system. This information is encoded as a bit vector, with each bit indicating whether a particular rule from a rule-based system would have fired in the context of the segment. The neural network replaces the action elements of a rule-based system, while still using the condition elements as input.

The hybrid network can also use a context window, using the rule firing information to supplement, rather than replace the other context information.

The rule firings used were similar to those described by Klatt [1], with some modifications. One input was used for each case of each rule. For example, the rule on postvocalic context of vowels has several cases, and each case is affected by the presence of phrase or clause boundaries. For each type of postvocalic context, there are two inputs: one which is used near phrase or clause boundaries, and one which is used elsewhere. This results in a total of 30 inputs for rule firings.

## 3. EMPIRICAL RESULTS

The database used to train the network consisted of 150 sentence recordings from a single speaker. The texts used for these recordings were a list of meaningful and semantically anomalous sentences which had been recorded for other purposes. An additional 10 sentences were reserved for testing the networks. These were labeled phonetically according to the guidelines used for the TIMIT database, and also had syllable, word, phrase, and clause boundaries marked. Syllable stress information and word type information was also marked.

A mean and standard deviation of the segment durations was computed for each type of phone. The neural network was trained to generate the number of standard deviations the duration of a segment differed from the mean for its phone.

A total of eight different networks were tested, with context window widths varying between one and seven, and each window width used both with and without the rule firing information. The results of these experiments are summarized in Figure 3, which shows the mean squared error in the neural network output when checked against the training data and the test data. The error is expressed as a percent of variance. Note that the error is the percent of the variance in the neural network targets. Much of the variance in the segment duration is explained by the phone expressed in the segment. This variation was eliminated by the scaling process, and was not present in the neural network targets.

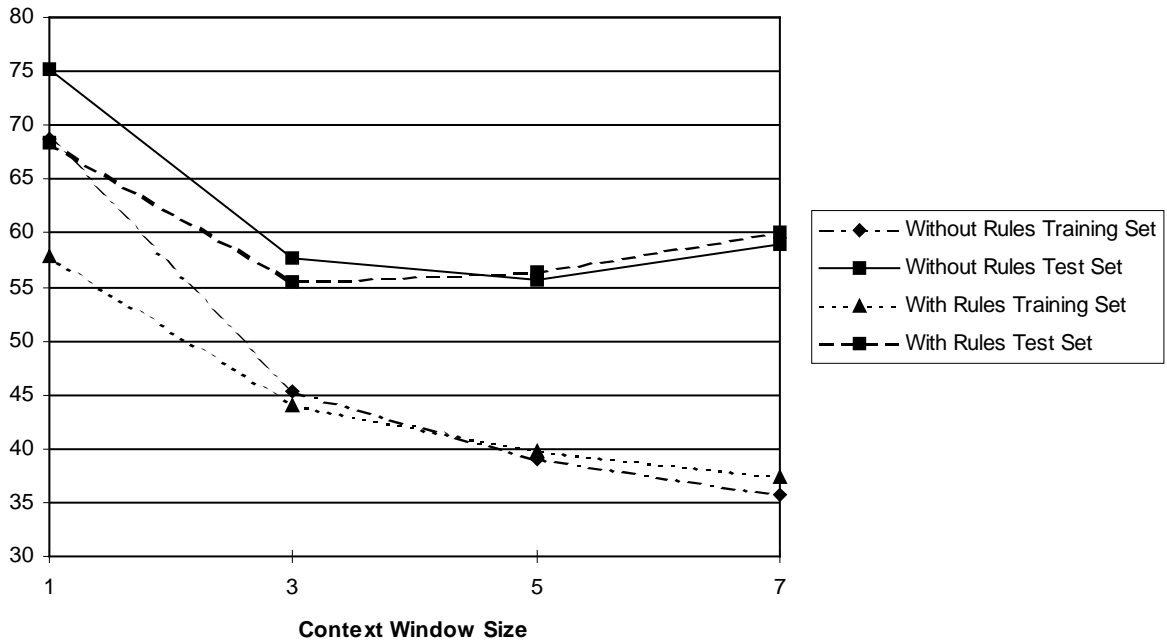

**Figure 3:** Mean square Error vs. Window Size

While the error observed in the training data fell steadily with the size of the context window, the error observed on the test data began increasing with window size after an initial drop. The rule firing information did help reduce the test error, although not as much as expected. On the other hand, the test error with a context window size of five without the rule firing information is greater than the test error for a context window size of three using the rule firing information, and the latter network is smaller, having a total of 14,291 weights, as compared to the network with a context window of 5, which had 21,715 weights. This represents a significant reduction in training time and execution time for the network using the rule firing information. Also, a context window size of seven would be necessary to model all of the situations detected by the rule firing information.

While it is true that with or without the rule firing information, the error observed on the test data never fell below fifty percent of the variance of the scaled segment duration, the error may not need to be reduced more. In an independent evaluation of the quality and intelligibility of the speech produced by various systems [4], speech synthesized using durations produced by this system was compared to speech synthesized using durations extracted from natural speech. The synthesis was performed using the neural network based system described in [3]. No significant difference was found between the performance of these two systems.

It is also useful to look at the effect each input has on the result produced by the neural network. Table 1 provides a rough estimate of this effect for a neural network using the rule firing information and a context window size of three phones. The first column lists the various types of binary inputs provided to the neural network. The second shows the sum of the absolute values of weights attached to inputs of that type in the first layer of the neural network. The value in the third column is the value in the second divided by the number of inputs of that type in the neural network, i.e., the average for a single input of the sum of the absolute value of the weights attached to an input of the type indicated in the first column. If an input has a large total, it would be expected to have a large effect on the neural network result. As the table shows, the rules have a significant effect. Only the syntax information, which marks boundaries that fall on particular phones, has a larger per-input contribution. It is also worth noting that three of the five rules with the greatest contribution are rules relating to syntactic boundaries. These are the rules that have a value of one for syllabic phones that are not in the final syllable of a word (total of weights: 12.1), for syllabic phones in a non-phrase final syllable (9.3), and for phones in the coda of a clause-final syllable (8.3). The other two of the five rules with the largest contributions relate to stress: an input that is one for phones in unstressed syllables (10.7), and one that is one for the nucleus of a syllable with secondary stress (9.6).

| Input Type | Absolute Weight Total | Mean Absolute Weight Total |
|---|---|---|
| Phone identity | 553.484 | 2.527 |
| Phone features | 758.500 | 4.770 |
| Syllable stress | 46.996 | 2.611 |
| Word Type | 118.805 | 1.331 |
| Syntax | 162.358 | 5.412 |
| Rules | 151.467 | 5.048 |

**Table 1:** Input Contributions

## 4. CONCLUSION

This paper describes the performance of a hybrid rule-based/neural network architecture for the computation of segment duration in a text-to-speech system. The hybrid exhibits slightly improved performance over a straight neural network system, but the improvement is less than expected. This may be adequate, however, as independent perceptual tests did not show a significant difference between speech produced with segment durations from this system and speech produced using segment durations extracted from natural speech. The hybrid network also involves less computation than a non-hybrid network with equivalent performance.